\documentclass[a4paper, 10pt, conference]{ieeeconf}      
\usepackage{FG2020}

\FGfinalcopy 

\IEEEoverridecommandlockouts                              
\usepackage{graphics} 
\usepackage{epsfig} 
\usepackage{mathptmx} 
\usepackage{times} 
\usepackage{amsmath} 
\usepackage{amssymb}  

\usepackage{float}
\usepackage{multicol}
\usepackage{multirow}
\usepackage{caption}
\usepackage{subcaption}

\def\FGPaperID{103} 

\title{\LARGE \bf
Pseudo-Convolutional Policy Gradient for Sequence-to-Sequence Lip-Reading
}

\author{\parbox{16cm}{\centering
    {\large Mingshuang Luo$^{1,2}$, Shuang Yang$^{1}$, Shiguang Shan$^{1,2}$} and Xilin Chen$^{1,2}$\\
    {\normalsize
    $^1$ Key Laboratory of Intelligent Information Processing of Chinese Academy of Sciences (CAS), Institute of Computing Technology, CAS, Beijing 100190, China\\
    $^2$ University of Chinese Academy of Sciences, Beijing 100049, China }}
}

\begin{document}

\ifFGfinal
\thispagestyle{empty}
\pagestyle{empty}
\else
\author{Anonymous FG2020 submission\\ Paper ID \103 \\}
\pagestyle{plain}
\fi
\maketitle

\begin{abstract}

Lip-reading aims to infer the speech content from the lip movement sequence and can be seen as a typical sequence-to-sequence (seq2seq) problem which translates the input image sequence of lip movements to the text sequence of the speech content.
However, the traditional learning process of seq2seq models always suffers from two problems: the exposure bias resulted from the strategy of ``teacher-forcing", and the inconsistency between the discriminative optimization target (usually the cross-entropy loss) and the final evaluation metric (usually the character/word error rate). 
In this paper, we propose a novel pseudo-convolutional policy gradient (PCPG) based method to address these two problems.
On the one hand, we introduce the evaluation metric (refers to the character error rate in this paper) as a form of reward to optimize the model together with the original discriminative target. 
On the other hand, inspired by the local perception property of convolutional operation, we perform a pseudo-convolutional operation on the reward and loss dimension, so as to take more context around each time step into account to generate a robust reward and loss for the whole optimization.
Finally, we perform a thorough comparison and evaluation on both the word-level and sentence-level benchmarks. The results show a significant improvement over other related methods, and report either a new state-of-the-art performance or a competitive accuracy on all these challenging benchmarks, which clearly proves the advantages of our approach.

\end{abstract}

\section{INTRODUCTION}
Lip-reading is an appealing tool for intelligent human-computer interaction and has gained increasing attention in recent years. It aims to infer the speech content by using the visual information \cite{B2017} like the lip movements, and so is robust to the ubiquitous acoustic noises. This appealing property makes it play an important role as the complement of the audio-based speech recognition system, especially in a noisy environment. At the same time, lip-reading is also crucial for several other potential applications, such as transcribing and re-dubbing archival silent films, sound-source localization, liveness verification and so on \cite{Chung}.
Benefiting from the vigorous development of deep learning (DL) and the emergence of several large-scale lip-reading datasets, such as GRID \cite{cooke2006}, LRW \cite{B2017}, LRW-1000 \cite{Yang2019}, LRS \cite{Chung} and so on, lip-reading has made great progress over the past two years.
	
\begin{figure}
	\setlength{\abovecaptionskip}{0.0cm}
	\setlength{\belowcaptionskip}{-0.2cm} 
	\centering
	\includegraphics[width=0.50\textwidth]{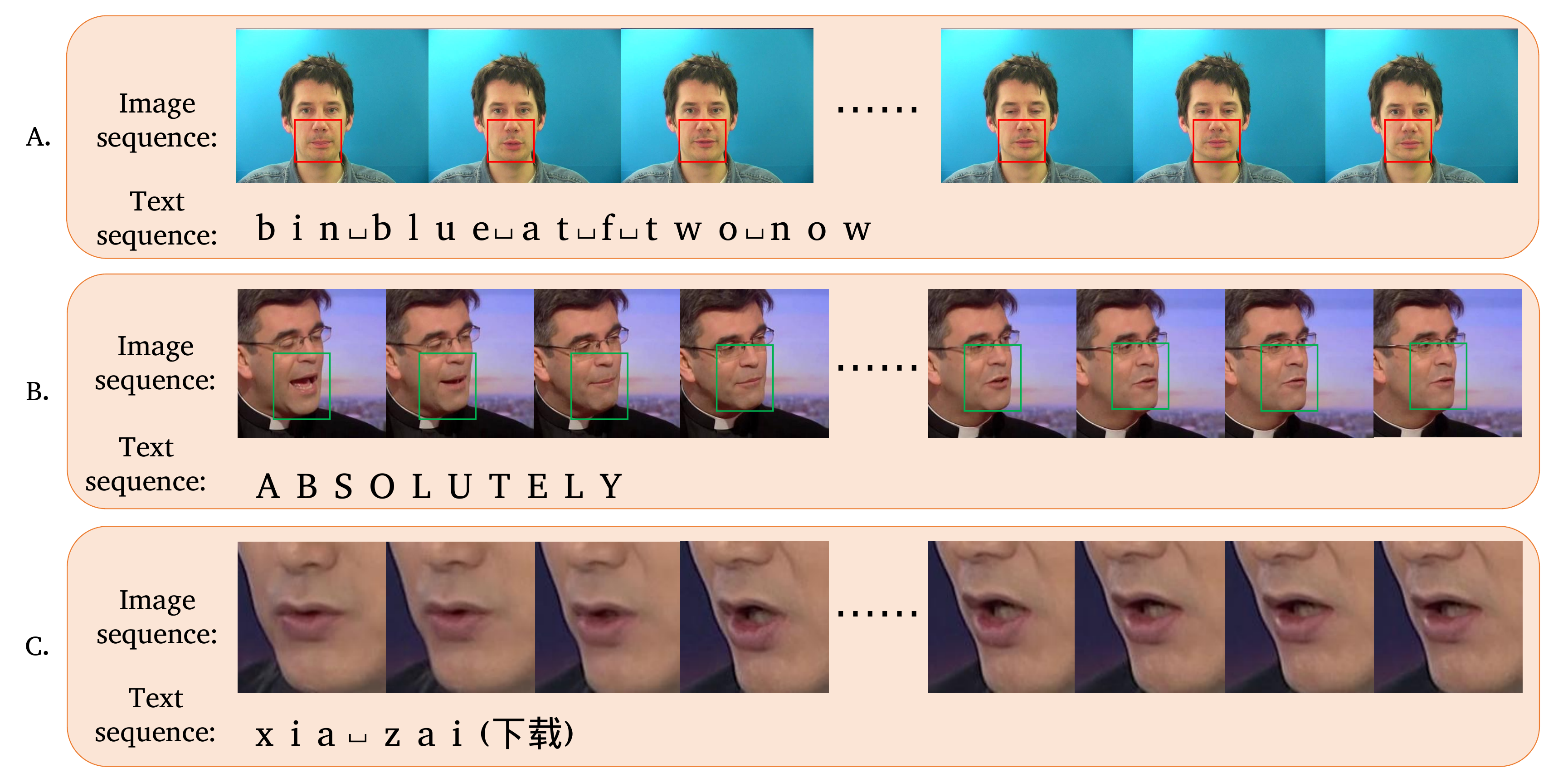}
	\caption{Examples for lip-reading. The examples in A-C are randomly sampled from the GRID, LRW and the LRW-1000 dataset, respectively.}
	\label{samples}
\end{figure}

Typically, lip-reading can be seen as a sequence-to-sequence (seq2seq) problem to translate the lip movement sequence to a character or word sequence, as shown in Fig. \ref{samples}. Several previous work have tried successfully to introduce seq2seq models for lip-reading \cite{Afouras2018}, \cite{Chung}, \cite{Afouras2017}, \cite{Chung2017}. 
However, most seq2seq-based methods suffer from two drawbacks. 
\textbf{The first problem} is the exposure bias resulted from the strategy of ``teacher-forcing''. Most current seq2seq models are learned in a way to obtain a correct prediction based on providing the ground-truth word at the previous time-step, which is named ``teacher-forcing” \cite{Rennie} and used widely in lip-reading \cite{Chung2017}, \cite{Chung}, \cite{Afouras2018}. This strategy is much favorable to make the model converge at a fast speed. But the optimized model learned in this way has a heavy dependence on the ground-truth words at the previous time-steps. Therefore, it is always difficult to obtain a consistent performance when using the optimized model for the actual test process, where no ground-truth is available and the model has to make a correct prediction based on the previous predictions \cite{Chopra2016}. This discrepancy would inevitably yield inaccurate results and even errors. Furthermore, this discrepancy would make the errors accumulate quickly along the sequence, leading to worse and worse predictions in the end. 
\textbf{The second problem} is the inconsistency between the optimized discriminative target and the final non-differentiable evaluation metric. For most lip-reading models, Cross-Entropy minimization (CE loss) is always applied as the discriminative optimization target. The CE loss is always used for each time step and the average over all time steps is finally used as the measure of the quality of the prediction results. This would always lead to two problems. Firstly, the optimized model may be not able to perform well when coming to test due to the inconsistency between the CE loss and the evaluation metric of WER/CER (word/character error rate). Secondly, the optimization process computes the cost at each time step independently, giving little consideration to the predictions before and after each time-step. This could probably lead to the case that only a single time step or just a few time steps are predicted well if they have a larger loss compared with the time steps nearby.
	
Inspired by the popular convolutional operation which has the appealing properties of local perception and weight sharing, we propose a novel pseudo convolutional policy gradient (PCPG) based seq2seq model to solve the above two problems for robust lip-reading. On the one hand, we introduce reinforcement learning (RL) into the seq2seq model to connect the optimized discriminative target and the evaluation metric of WER/CER directly. On the other hand, we mimic the computation process of traditional convolutional operation to consider more time steps nearby when computing the reward and loss at each time step. By a thorough evaluation and comparison on several lip-reading benchmarks, we demonstrate both the effectiveness and the generalization ability of our proposed model. Finally, we report either new state-of-the-art performance or competitive accuracy on both the word-level and sentence-level benchmarks.

\section{Related work}
In this section, we firstly give a brief review of the previous work for the lip-reading task. Then we discuss the work related with seq2seq models and reinforcement learning involved in our model.

\subsection{Lip-reading}
Lip-reading has made several substantial progress since the emergence of deep learning (DL) \cite{Chung2018, Chung, Chung2017, B2017, Petridis2018, Assael2016, Afouras, Zhou2014, Yang2019, Hu2016}. These contempts can be divided into two strands according to their particular task. 

The first one mainly focuses on word-level lip-reading. This type of method aims to classify the whole image sequence into a single word class, no matter how many frames in the sequence. For example, the work in \cite{Chung2017} proposed to extract frame-level features by the VGG network and then combine all the features of all the frames at several different stages to obtain the final prediction of the whole sequence. 
In \cite{Stafylakis2017},  the authors proposed an end-to-end deep learning architecture by combining a spatiotemporal convolutional module, the ResNet module and the final bidirectional LSTM for word-level lipreading, and have obtained the state-of-the-art performance when they were published. In this paper, we would keep a similar front-end module with them and would present details in the next section. 

The second strand pays more attention to sentence-level lip-reading task. Different from the word-level which considers the whole image sequence as a single category, the sentence-level method has to decode the character or word at each time step to output a sentence. The LipNet model presented in \cite{Assael2016} is the first end-to-end sentence-level lip-reading model. It employes the CTC loss to predict different characters at different time steps, which would then compose the final sentence-level predictions. Then, the work in \cite{Afouras} evaluated not only the CTC loss but also the beam search based process \cite{Wiseman2016} for lip-reading. However, the CTC based methods are all based on the assumption that all the predictions at each time step are independent of each other, which we think is not proper in a sequence-based task. In the meantime, the CTC loss has also a limitation that the length of the input image sequence has to be larger than the length of the speech content sequence. Therefore, seq2seq models are gradually introduced to the sentence-level lip-reading.

\subsection{Seq2Seq models}
Sequence-to-sequence (seq2seq) models always contain an RNN-based encoder to encode the input image sequence into a vector and an RNN-based decoder to generate the prediction result at each time step. 
In the decoding process, attention mechanism is always introduced to make the prediction operation at each output's time step being able to ``observe'' all the input sequence's time steps. 
For example, \cite{Chung, Chung2017, Afouras2018} have successfully introduced seq2seq models for sentence-level lip-reading.

However, the usual seq2seq models always suffer from two main limitations.
Firstly, most current seq2seq models are trained in the way of ``teacher forcing'', which would provide the ground-truth word at the previous time step as a condition to predict the results at the next time step. In fact, the models have to predict each output based on the previous prediction results when comes to testing, where no ground-truth words are available.
This could easily lead to error accumulation along with the sentence, which is not the result we want.
Secondly, most seq2seq models are optimized by minimizing the sum of the cross-entropy loss at each time step. But when it comes to evaluation, the metrics take the whole sentence into account with discrete and non-differentiable metrics, such as BLEU \cite{Papineni2002}, ROUGE \cite{Lin2001}, METEOR \cite{Banerjee2003}, CIDEr \cite{Tech}, WER (word error rate), CER (character error rate), SER (sentence error rate), and so on. The inconsistency between the loss and the evaluation metric is a serious problem, which could easily lead to the case that even we get a very small cross-entropy loss during training, we may still not be able to get a good performance when it comes to testing. 

\begin{figure*}[htb]
	\setlength{\abovecaptionskip}{0.2cm}
	\setlength{\belowcaptionskip}{-0.3cm} 
	\centering
	\includegraphics[width=1.00\textwidth]{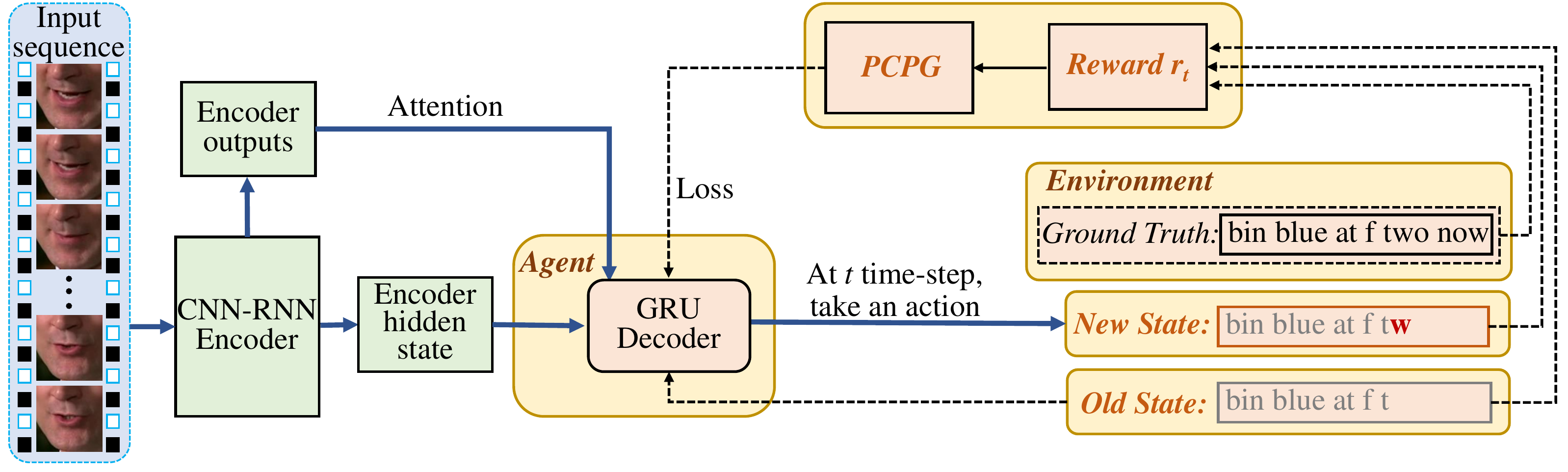}
	\caption{Overview of our proposed PCPG based seq2seq model for lip-reading. In this model, the GRU decoder is regarded as an agent and the ground truth is regarded as the environment. In the learning process, the agent observes the old state generated from the previous time steps and then takes an action to output a new character/word to obtain a new state. The new state, old state, and the environment would contribute to the reward together for the action. Finally, the reward is fed to the PCPG module to generate the final loss when passed to the agent.}\label{overview_fig}
\end{figure*}

In this paper, we propose a novel pseudo-convolutional policy gradient (PCPG) based seq2seq models for the lip-reading task, as shown in Fig. \ref{overview_fig}. 
In our model, the evaluation metric is introduced directly as a form of reward to optimize the model together with the original discriminative target.
At the same time, a pseudo-convolutional operation is performed on the reward and loss dimension to take more context around each time step into account to generate a robust reward and loss for the whole optimization. The pseudo-convolutional operation is to mimic the local perception and weight sharing property of convolutional operation, contributing to enhancing the contextual relevance among different time steps for robust lip-reading.

\section{The Proposed work}
\label{sec:Pro}
In this section, we present the proposed PCPG (Pseudo-Convolutional Policy Gradient) based seq2seq model in detail. Firstly, we describe the overall model architecture in the first subsection. Then we introduce the PCPG based learning process. Finally, we state the advantages of our method compared with the traditional methods.

\subsection{The Overall Model architecture} \label{section3.1}
As shown in Fig. \ref{overview_fig}, our model can be divided into two main parts: the video encoder (shown with green color) and the GRU based decoder (shown with yellow color). The video encoder is responsible for encoding the spatiotemporal patterns in the image sequence to obtain a preliminary representation of the sequence. After encoding, the GRU decoder tries to generate predictions at each time step in the principle of making the reward at each time step maximized.

\begin{figure}[htb]
	\setlength{\abovecaptionskip}{0.2cm}
	\setlength{\belowcaptionskip}{-0.2cm} 
	\centering
	\includegraphics[width=0.48\textwidth]{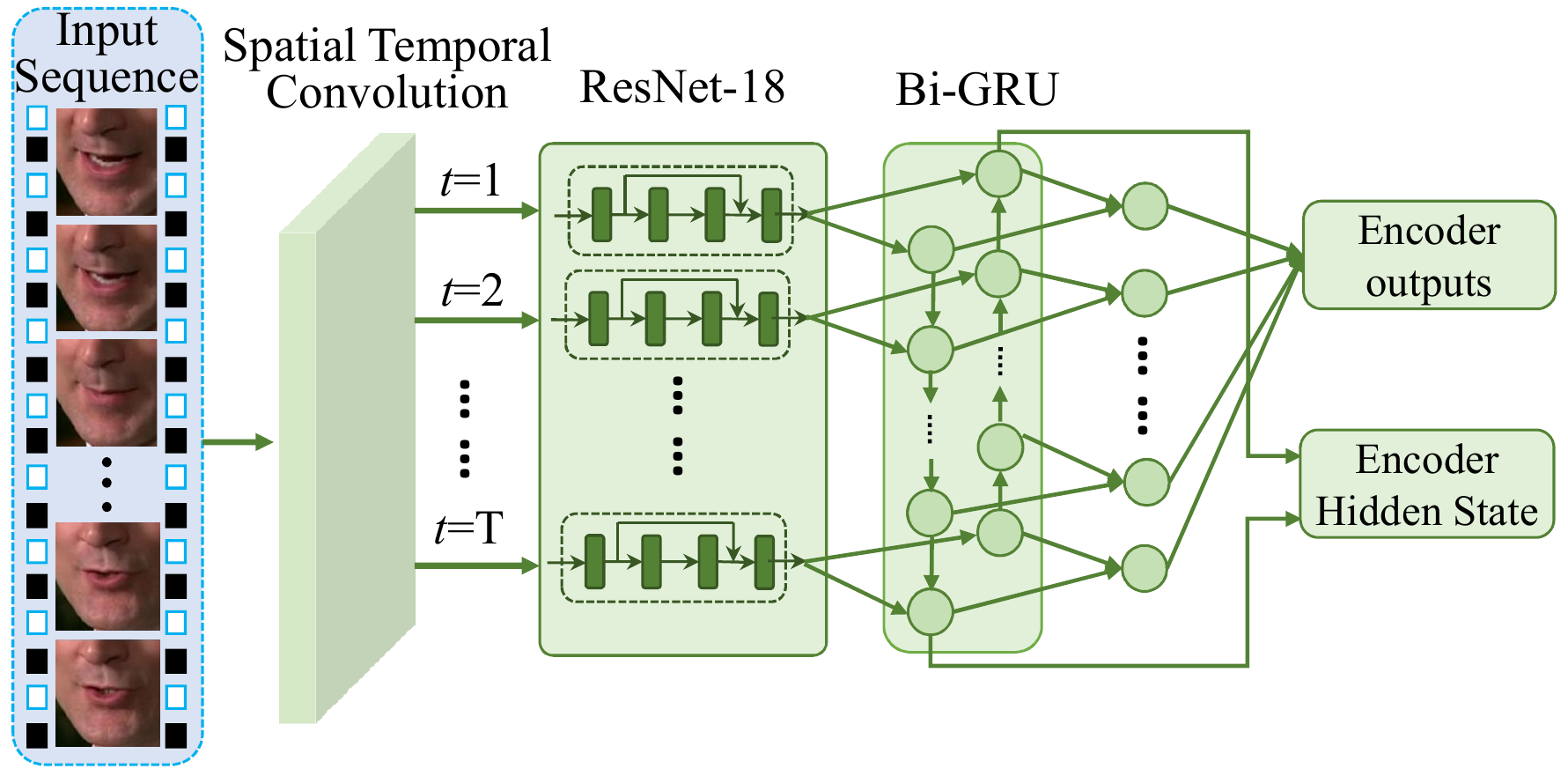}
	\caption{The CNN and RNN based video encoder. }\label{vid_encoder_fig}%
\end{figure}
\vspace{0.2cm}
\indent{\bf The CNN and RNN based Encoder}: As shown in Fig. \ref{vid_encoder_fig}, the encoder consists of two main modules: the CNN based front-end to encode the short-term spatial-temporal patterns, and the RNN based back-end to model the long-term global spatial-temporal patterns. The input image sequence would firstly go through a 3D Convolutional layer to capture the initial short-term patterns in the sequence. Then a ResNet-18 \cite{He} module is followed to capture the specific movement patterns at each time step. Finally, a 2-layer Bi-GRU is adopted to produce a global representation of the whole sequence. We use the GRU's output $\textit{\textbf{O}}^{e}=\left(\mathbf{o}_{1}^{e}, \mathbf{o}_{2}^{e}, \dots, \mathbf{o}_{T}^{e}\right)$ and hidden-state vector $\textit{\textbf{h}}^{e}$ to record the patterns of the input video $\mathbf{X}^{v}=\left(\mathbf{x}_{1}^{v}, \mathbf{x}_{2}^{v}, \dots, \mathbf{x}_{T}^{v}\right)$, which is defined as:
$$
\textit{\textbf{h}}^{e}, \textit{\textbf{O}}^{e}={Encoder\_CNN\_RNN}(\mathbf{X}^{v}), \eqno{(1)}
$$
where $e, v, T$ denote the \textit{encoder}, the original input \textit{video} and the \textit{temporal length} of input video rerspectively.

\begin{figure*}
	\setlength{\abovecaptionskip}{0.05cm}
	\setlength{\belowcaptionskip}{-0.3cm} 
	\centering
	\includegraphics[width=0.98\textwidth]{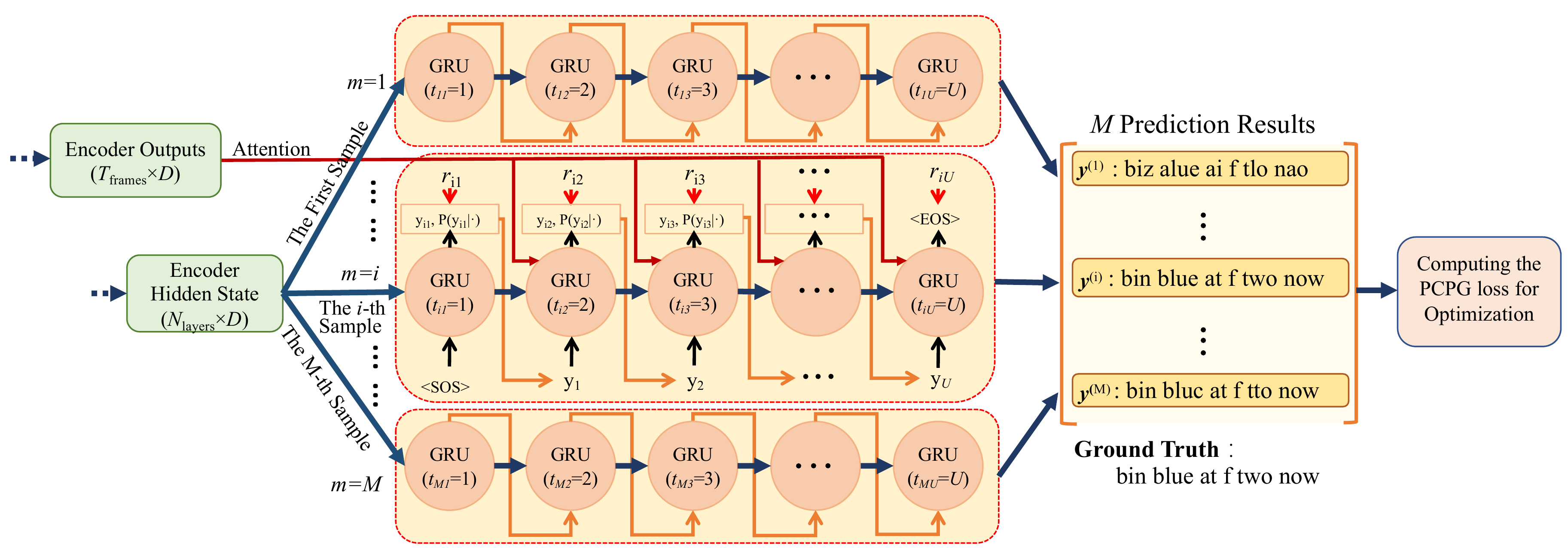}
	\caption{The Optimization of PCPG based learning process. Each output at the previous time step will feedback to the agent (GRU decoder) as the input. And the environment will feedback on an immediate reward when taking an action (choosing a character at each time step). To calculate the gradient for model updating, we utilize Monte Carlo sampling to sample M transcription sequences, leading to M output character sequences which would be taken to compute the PCPG loss.}\label{decoder_with_RL}
\end{figure*} 

\indent{\bf The PCPG based RNN Decoder}: Our PCPG based RNN decoder is showed as Fig. \ref{decoder_with_RL}. Given the representation of each input sequence, a 2-layer RNN is followed to decode each character at each output's time step $u~(u=1,2,..., U)$, where $U$ denote the maximum length of the output text sequence. The PCPG based loss would then guide the learning process of the model. In this paper, we use GRU as the basic RNN unit. 

To learn the dependency of each output character with each input time step $t$ $(t=1,2,...,T)$, attention mechanism is introduced into the decoding process, which takes advantage of the context information around each time step $t$ to aid the decoding of each output's time step $u$, where $t$ and $u$ are used to denote the time steps in the input image sequence and the output text sequence respectively. With the decoder GRU's hidden state $h_{u-1}^{d}$ $(u=1,2,...,U)$ where $d$ denotes \textit{decoder}, we can compute the attention weight $\alpha_{u}$ on the current time step $u$ with respect to each input time step $t$ and the corresponding output $y_u$ as 
$$ 
\begin{array}
{c}{{e}_{u, t}=attention\left(\boldsymbol{h}_{u-1}^{d},~{o}_{t}^{v}\right) }, \\[2mm] 
{{a}_{u, t}=\frac{e_{u,t}}{\sum_{t} e_{u,u}}}, \quad {\alpha_{u}=\sum_{t} a_{u t} o_{t}^{v}}, \\[2mm]
\boldsymbol{y}_{u}={GRU}\left(\boldsymbol{h}_{u-1}^{d}, \boldsymbol{y}_{u}^{d},{a_{u}}\right). 
\end{array}
\eqno{(2)}
$$
where $t=1, 2, ..., T$ and $u=1, 2, ..., U$. Finally, $(y_1, y_2, ..., y_U)$ would be used as the final output.

\subsection{The PCPG based learning process }
With the model given above, a popular way to learn the model is to minimize the cross-entropy loss  $L_{CE}$ at each time step as follows:
$$
{L}_{CE}=-\log p\left(c_{1}, c_{2}, \ldots, c_{U}\right)
$$
\vspace{-0.5cm}
$$
~~~~~~~~~~~~~=-\sum_{u=1}^{U} \log p\left(c_{u} | c_{1}, c_{2}, \ldots, c_{u-1}\right).     
\eqno{(3)}
$$
where $c_{u} = 1, 2, ..., C$ is the predicted class label index at time step $u$, $C$ is the number of categories to be predicted at each time step. In this paper, there are $C = 40$ categories at each time step $u$, including $26$ alphabets, $10$ numbers, the space, the begining, padding and ending symbol. 

At each time step $u$, the model would generate a prediction result according to the predictions at previous time steps $1, 2, ..., u-1$. In other words, the prediction at each time step would be decided by the predictions at the previous time step.  

Besides the above optimization target, we also view the seq2seq model as an `agent' in this paper that interacts with an external `environment' which is corresponding to video frames, words or sentences here, as shown in Fig. \ref{overview_fig} with the parameters of the model denoted as $\theta$, the model can be viewed as a policy $p_{\theta}$ leading to an `action' of choosing a character to output. At each time step $u$, the agent will get a new internal `state', which is decided by the attention weight $a_{u}$, the previous hidden state $h_{u-1}$ and the predicted character $y_{u}$. With this state, there would be an immediate reward $r_{u}$ to evaluate the reward and cost of predicting the character $y_u$ at the time step $u$. Then the training goal is to maximize the expected reward $E_{y}[R|p_{\theta}]$ where $R$ refers to the cumulative reward. In the following, we would describe the reward function and the principle to update the parameters in this paper in detail.

\indent{\bf Reward function}: In lip-reading tasks, the performance of the model is finally evaluated by CER or WER, both of which are usually obtained by the edit-distance or Levenshtein distance between the predicted word/sentence and the ground truth word/sentence. Here, we choose the negative CER of the whole sentence as the immediate reward $r_{u}$ to evalute the effect of the prediction at each time step $u$, which is defined as follows:
$$ 
r_{u}=\left\{\begin{array}{ll}{-\left(E D\left(\mathbf{y}_{1 : u}, S\right)-E D\left(\mathbf{y}_{1 : u-1}, S\right)\right)} & {\text { if } u>1} \\ {-\left(E D\left(\mathbf{y}_{1 : u}, S\right)-\left|S\right|\right)} & {\text { if } u=1.}\end{array}\right.
\eqno{(4)}
$$
where $S$ refers to the ground truth text sequence and $|S|$ is the length of $S$, $ED(a, b)$ refers to the CER between characeter sequences $a$ and $b$, which is computed by the edit-distance.

An example of the process is shown in Fig. \ref{overview_fig}, where the ground-truth $S$ is the sequence of \textit{`bin blue at f two now'}, the old state $\mathbf{y}_{1 : u-1}$ (i.e. the previous decoding sequence) is \textit{`bin blue at f t'}. The model observes the old state and then takes an action of choosing a character \textit{`w'} to generate a new state $\mathbf{y}_{1 : u}$, corresponding to \textit{`bin blue at f tw'}. We would compute the reward $r_{u}$ at the time step $u$ as Eq. (4) to evaluate the reward of predict \textit{`w'}.

\begin{figure*}
	\centering
	\setlength{\abovecaptionskip}{-0.50cm}   
	\setlength{\belowcaptionskip}{-0.50cm} 
	\includegraphics[width=0.9  \textwidth]{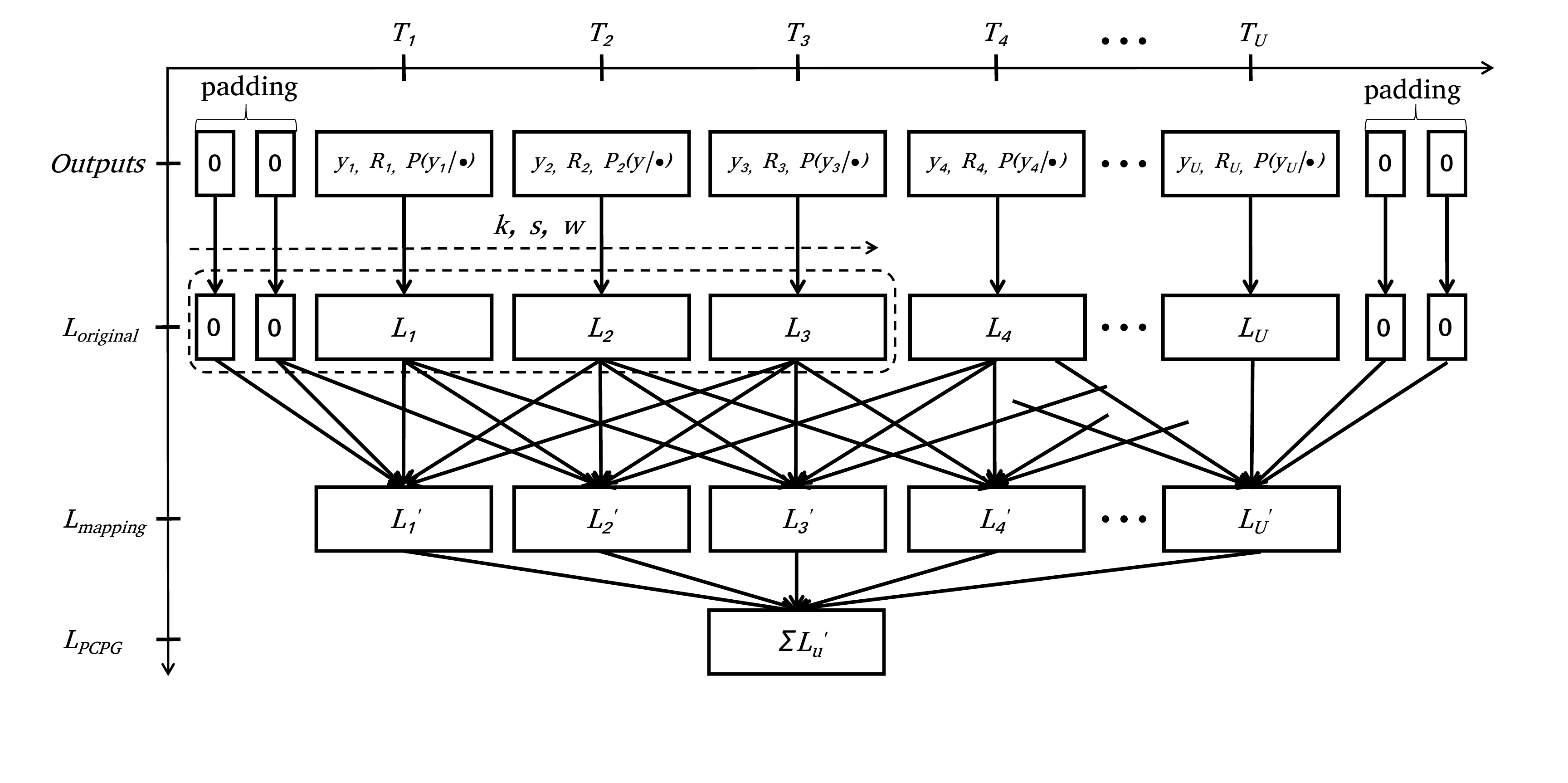}
	\caption{The computation process of the PCPG based loss function, where the kernel size $k$ is 5, the stride $s$ is 1, and the kernel weights $w$ is [$\dfrac{1}{5}$, $\dfrac{1}{5}$, $\dfrac{1}{5}$, $\dfrac{1}{5}$, $\dfrac{1}{5}$] in this figure.} \label{pcpg}
\end{figure*}

\indent{\bf Optimization}: We use $r_u$ and $R_{u}$ to denote the immediate reward at time step $u$ and the future expected reward at time step $u$ respectively.
Given the reward $r_u$ at each time step $u$, the future expected reward $R_{u}$ at current time step $u$ would be computed as $R_{u}=\sum_{i=u}^{U}\gamma^{U-i}r_{i}$, where $\gamma$ is the discount factor and $U$ is the max length of the character sequence.  The final reward for the whole sequence $R$ is $R=\sum_{u=1}^{U}R_{u}$, which is used to denote the cumulative reward of the whole prediction result from the begining to  the end $U$ ($\mathbf{y}_{1 : U}$). 
Inspired by the properties of local perception and weight sharing of convolutional operation, we compute the PCPG based loss $L_{PCPG}$ as
$$
L_{u}=L_{original}=-R_{u}\cdot
logP(y_{u}|\cdot;\theta). \eqno{(5)}
$$
$$
L_{u}^{'}=L_{mapping}=L_{u-|k/2|:u+|k/2|}\cdot w=\sum_{i=u-|k/2|}^{u+|k/2|}w_{i}\cdot
L_{u}. \eqno {(6)}
$$
$$
L_{PCPG}= \sum_{u=1}^{U}L_{u}^{'}=\sum_{i=u-|k/2|}^{u+|k/2|}\sum_{u=1}^{U}w_{i}\cdot
L_{i}. \eqno{(7)}
$$ 
where $w$, $k$, $\theta$, $p_{\theta}$ denotes the kernel weights, the kernel size, the parameters' distribution of the model and the parameters. The computation process is shown in Fig. \ref{pcpg} (where $k=5, s=1, w=[1/5, 1/5, 1/5, 1/5, 1/5]$). To ensure the value of the gradient in PCPG at the same quantitative level as the traditional PG, we set $\sum_{i=1}^{k}w_{i}=1$ in our paper where $k$ is the kernel size and $w_{i}$ is the kernel weight.

Finally, when a pair of prediction result is generated, and denoted as $\mathbf{X}^{v}=\left(\mathbf{x}_{1}^{v}, \mathbf{x}_{2}^{v}, \dots, \mathbf{x}_{T}^{v}\right)$, $\mathbf{y}=\left(y_{1}, y_{2}, \dots, y_{U}\right)$, we would use $L_{combine}$ as our final loss for our PCPG based seq2seq model which $L_{combine}$ can be defined as:
$$L_{combine}=(1-\lambda)\cdot L_{CE} + \lambda\cdot L_{PCPG}, \eqno{(8)}$$
where the $\lambda$ is a scalar weight to balance the two loss functions.

In practice, it is always much difficult to integrate all possible transcriptions ($\mathbf{y}$) to compute the above gradient of the expected reward. So we introduce Monte Carlo sampling here to sample $M$ transcription sequences $\mathbf{y}^{(1)}, \mathbf{y}^{(2)}, ..., \mathbf{y}^{(M)}$ to estimate the true gradient, which is shown in Fig. \ref{decoder_with_RL}. So the gradient can be computed finally as:
$$
\nabla_{CE-\theta}=-\frac{1}{M}\sum_{m=1}^{M}\sum_{u=1}^{U_{(m)}}\nabla_{\theta}\log p\left(y_{u}^{(m)} | y_{1}^{(m)}, y_{2}^{(m)}, \ldots, y_{u-1}^{(m)};\theta\right). \eqno{(9)}
$$
$$\nabla_{PCPG-\theta}
\approx-\frac{1}{M} \sum_{m=1}^{M} \sum_{u=1}^{U_{(m)}} \sum_{i=u-|k/2|}^{u+|k/2|}w_{i}\cdot
R_{u}^{(m)} \nabla_{\theta} \log P\left(y_{u}^{(m)} | \mathbf{y}_{<u}^{(m)}, \mathbf{X}^{v} ; \theta\right).
\eqno{(10)}
$$
Therefore,  the parameters could be updated as follows: 
$$ 
\frac{\partial L_{combine}(\theta)}{\partial \theta} \approx(1-\lambda)\cdot \nabla_{CE-\theta} + \lambda\cdot \nabla_{PCPG-\theta}
$$
$$
\theta^{'}=\frac{\partial L_{combine}(\theta)}{\partial \theta}\cdot lr+\theta
\eqno{(11)}
$$
where $lr$ denotes the learning rate.

\subsection{Compared with traditional Policy Gradient}   \label{section3.4}
Compared with the traditional policy gradient (PG), the PCPG has two more important operations as the convolutional operation: \textbf{local perception} and \textbf{weights sharing} shown as Fig. \ref{pcpg}. In traditional PG, there is no concept of receptive field and the loss function could not make use of the context information. However, the lip-reading task can be considered as a sequence-to-sequence task and, the context is very important for accurate decoding results. With the existence of local perception and weights sharing, the prposed PCPG can make the proposed model to establish stronger semantic relationships among different time steps in the optimization process.

On the other hand, it is well-known that it is usually unstable for the models to train with RL \cite{Berkenkamp2017, Nikishin2018, Jin2019}. There will be a big gradient change due to the randomness in the process of deciding an action with a traditional PG algorithm. While in our work, we can find that the immediate loss $L_{u}$ at each time-step will be used to generate an average value based on multiple time-steps in PCPG, as shown in Eq.(7). 
At the same time, the existence of the overlapping parts has further made the local gradient value not change dramatically. 
So the model can obtain more favorable contextual loss constraints to make the convergence more stable and faster with our proposed PCPG.


\section{Experiments}  \label{section4}
In this section, we evaluate our method on three large-scale benchmarks, including both the word-level and sentence-level lip-reading benchmarks. At the same time, we also discuss the effects of the pseudo-convolutional kernel's hyperparameters $k$, $w$, $s$ on the performance of lip-reading through a detailed ablation study. By comparing with several other related methods, the advantages of our proposed PCPG based seq2seq model are clearly shown.
\subsection{Datasets}
We evaluate our method on three datasets in total, including the sentence-level benchmark, GRID,  and the large-scale word-level datasets, LRW and LRW-1000.

{\bf GRID \cite{cooke2006}}, released in 2006, is a widely used sentence-level benchmark for the lip-reading task \cite{Assael2016}, \cite{Lan2009}, \cite{Wand2016}. There are 34 speakers and each one speaks out 1000 sentences, leading to about 34,000 sentence-level videos in total. All the videos are recorded with a fixed clean single-colored background and the speakers are required to face the camera with the frontal view in the speaking process. 

{\bf LRW \cite{B2017}}, released in 2016, is the first large scale word-level lip-reading datasets. The videos are all collected from BBC TV broadcasts including several different TV shows, leading to various types of speaking conditions in the wild. Most current methods perform word-level tasks using classification-based methods. In our experiments, we try to explore the potential of seq2seq models for the word-level tasks but we also perform classification based experiments to evaluate the representation learned by our PCPG based model.

{\bf LRW-1000 \cite{Yang2019}}, released in 2018, is a naturally-distributed large-scale benchmark for Mandarin word-level lip-reading. There are 1000 Mandarin words and more than 700 thousand samples in total. Besides owning a diversified range of speakers' pose, age, make-up, gender and so on, this dataset has no length or frequency constraints in the words, forcing the corresponding model to be robust and adaptive enough to the practical case where some words are indeed more or less longer or frequent than others. These properties make LRW-1000 very challenging for most lip-reading methods. 
\vspace{0.3cm}
\subsection{Implementation details}
In our experiments, all the images are normalized with the overall mean and variance of the whole dataset. When fed into models, each frame is randomly cropped, but all the frames in a sequence would be cropped in the same random position for training. All frames are centrally cropped for validation and test.  

Our implementation is based on PyTorch and the model is trained on servers with four NVIDIA Titan X GPUs, with 12GB memory of each one. We use Adam optimizer with an initial learning rate of 0.001. Dropout with probability 0.5 is applied to the RNN and the final FC layer of the model. 
For PCPG, we consider the following three situations: (1) $k$=1, $s$=1 which is degenerated to the usual REINFORCEMENT algorithm, i.e. the traditional PG algorithm in this case. (2) $k=5$, $s=5$, which is sample PCPG version without overlapping parts. (3) $k=5$, $s=1$, which has 4-time steps as the overlapped part between twice adjacent computation. Here, we use CER and WER as our evaluation metrics. The $\uparrow$ denotes that larger is better while the $\downarrow$ denotes that lower is better. And the kernel's weight $w$ is set to [1/5, 1/5, 1/5, 1/5, 1/5] by default.

In this paper, we would use the common cross-entropy loss $L_{CE}$ (shown as Eq. (3)) based seq2seq model as our baseline model.
To evaluate the representation generated by our PCPG based method, we also perform classification based experiments on the LRW and LRW-1000 with a fully connected (FC) classifier based back-end. Specifically, we firstly trained the PCPG based seq2seq model in LRW and LRW-1000 with the loss $L_{combine}$ and obtained a video encoder and a GRU based decoder. Then we just fixed the encoder part to fix the representation learned by the PCPG based seq2seq model and then trained the FC based classifier with this representation, where the loss used for the FC classifier $L_{Classify}$ is defined as:
$$
L_{Classify}=-\sum_{k}{P}_{k}\log{P}_{k}.
\eqno{(12)}
$$
\begin{figure*}
	\centering
	\setlength{\abovecaptionskip}{-0.00cm}   
	\setlength{\belowcaptionskip}{-0.00cm} 
	\includegraphics[width=0.99 \textwidth]{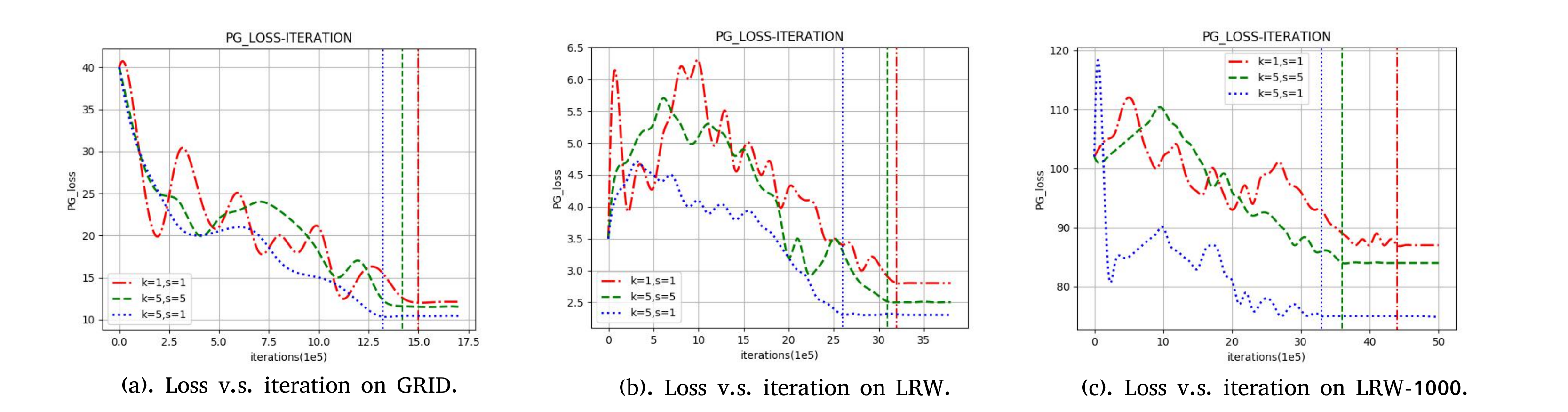}
	\caption{loss v.s. iteration on GRID, LRW, LRW-1000. The three vertical lines (blue, green, red) in (a), (b) and (c) refer to the state when the model converges under different conditions.} \label{figure6}
\end{figure*}
\vspace{-0.6cm}
\subsection{Ablation Study}
By performing like the convolutional operation, the proposed PCPG has gained the propertis of receptive field ($\mathbf{RF}$) and overlapping parts ($\mathbf{OP}$). When there is no overlapping parts and the receptive field is equal to 1, the proposed PCPG based seq2seq just equals to the traditional PG based seq2seq.  We summarize the detailed comparison of different choices of $\mathbf{RF}$ and $\mathbf{OP}$ in TABLE \ref{table1}, where `$+ \mathbf{RF}$' and `$- \mathbf{RF}$' means $\mathbf{RF}>1 (k>1)$ and $\mathbf{RF} =1 (k=1)$,  and `$+ \mathbf{OP}$' and `$- \mathbf{OP}$' means $\mathbf{OP}>1 (k>s)$ and $\mathbf{OP} =1 (k=s)$ respectively. $k$ and $s$ is set to 5 by default when they are not 1. 

From TABLE \ref{table1}, we can see that the model with traditional RL when there is no extra \textbf{RF} and \textbf{OP} ($k=1, s=1$) performs better than the traditional cross-entropy based seq2seq baseline. This shows that RL is an effective strategy to improve the seq2seq model for lip-reading.
When both \textbf{RF} and \textbf{OP} are effective ($k=5, s=1$), the best performance is given, which proves the effectiveness of the proposed PCPG for robust lip reading. Besides the comparison with the baseline, we also present the loss curves in different settings during the training process on different benchmarks in Fig. \ref{figure6} (a), (b), (c). We can see that the PCPG can make the model learned more stably and also take less time to converge than the other two baselines.
\vspace{-0.16cm}

\begin{table}[H]
	\setlength{\abovecaptionskip}{0.0cm}
	
	\caption{The results of the seq2seq model with different settings on three lip-reading datasets. (\textbf{RF}: receptive field, \textbf{OP}: overlapping parts)}
	\label{table1}
	\begin{center}
		\begin{tabular}{|c||c|c||c|r|}    
			\hline
			Dataset & Method& Case& CER $\downarrow$ &  WER $\downarrow$ \\
			\hline
			\hline
			\multirow{4}{*}{GRID}&    $L_{CE} (baseline)$ & / &8.4$\%$ & 18.3$\%$ \\
			~&    $-$ \textbf{RF} and $-$ \textbf{OP}&$k$=1, $s$=1& 7.6$\%$ & 16.6$\%$\\
			~&    $+$ \textbf{RF} and $-$ \textbf{OP}&$k$=5, $s$=5& 6.9$\%$ & 15.3$\%$ \\
			~    &$+$ \textbf{RF} and $+$ \textbf{OP} &$k$=5, $s$=1& \textbf{5.9}$\%$ & \textbf{12.3}$\%$ \\
			\hline
			\multirow{4}{*}{LRW}&$L_{CE} (baseline)$ & /&17.5$\%$ & 28.3$\%$  \\
			~& $-$\textbf{RF} and$-$ \textbf{OP}&$k$=1, $s$=1& 15.2$\%$ & 24.8$\%$ \\
			~&$+$ \textbf{RF} and $-$ \textbf{OP}&$k$=5, $s$=5& 15.0$\%$ & 26.5$\%$ \\
			~&$+$ \textbf{RF} and $+$ \textbf{OP} &$k$=5, $s$=1& \textbf{14.1}$\%$ & \textbf{22.7}$\%$ \\
			\hline
			\multirow{4}{*}{LRW-1000}&$L_{CE} (baseline)$             &/  & 52.1$\%$ & 68.2$\%$     \\
			~&$-$ \textbf{RF} and $-$ \textbf{OP} &$k$=1, $s$=1& 51.4$\%$ & 67.7$\%$    \\
			~&$+$ \textbf{RF} and $-$ \textbf{OP} & $k$=5, $s$=5& 51.6$\%$   &   67.2$\%$      \\
			~&$+$ \textbf{RF} and $+$ \textbf{OP} &$k$=5, $s$=1& \textbf{51.3}$\%$ & \textbf{66.9}$\%$      \\
			\hline
		\end{tabular}
	\end{center}
\end{table}


\begin{table}[H]    
	\setlength{\abovecaptionskip}{0.0cm}
	\caption{Evaluation of parameters $k$ and $w$. (a) results with different $k$ ($s=1$); (b) results with different choices of $w$ ($k$=3, $s$=1).}\label{table2}
	\centering
	\begin{subtable}[t]{1in}
		\centering
		\begin{tabular}{|c||c|}    
			\hline
			Kernel size\centering      &     WER $\downarrow$  \\
			\hline
			\hline
			$k$=1& 16.6$\%$\\
			$k$=2 &16.0$\%$ \\
			$k$=3&\textbf{12.1}$\%$ \\
			$k$=5 & 12.3$\%$ \\  
			$k$=7&14.8$\%$ \\
			\hline
		\end{tabular}
		\caption{Evaluation on $k$}\label{table2:a}
	\end{subtable}
	\quad
	\begin{subtable}[t]{2in}
		\centering
		\begin{tabular}{|c||c|}    
			\hline
			Kernel weight&     WER $\downarrow$  \\
			\hline
			\hline
			$w$=[1/3,1/3,1/3]& 12.1$\%$\\
			$w$=[1/4,1/2,1/4]&\textbf{11.9$\%$} \\
			$w$=[1/3,1/2,1/6]&12.7$\%$ \\
			$w$=[1/6,1/2,1/3]&12.6$\%$ \\  
			\hline
		\end{tabular}
		\caption{Evaluation on $w$}\label{table2:b}
	\end{subtable}
\end{table}    

\vspace{-0.7cm}
\begin{table}[H]
	\setlength{\abovecaptionskip}{0cm}   
	\setlength{\belowcaptionskip}{0cm} 
	\centering
	\caption{Evaluation of the learned representation with classification based back-end on LRW and LRW-1000. (\textbf{baseline}: without any extra pretraining, \textbf{FE}: fix encoder, \textbf{TE}: train encoder, \textbf{TC}: train classifier.) 
		\vspace{-0.0cm}} \label{table4}
	\begin{tabular}{|p{1.39cm}|c|c|ccr}    
		\hline
		Dataset & Method&  Accuracy $\uparrow$\\
		\hline
		\hline
		\multirow{3}{*}{LRW}&\textbf{baseline} & 82.1$\%$  \\
		~&\textbf{FE} and \textbf{TC}&  82.4$\%$ \\
		~&\textbf{TE} and \textbf{TC} & \textbf{83.5}$\%$  \\
		\hline
		\multirow{3}{*}{LRW-1000}&\textbf{baseline} & 37.8$\%$      \\
		~&\textbf{FE} and \textbf{TC} &  38.5$\%$     \\
		~&\textbf{TE} and \textbf{TC} &\textbf{38.7}$\%$      \\
		\hline
	\end{tabular}
	
\end{table}

\subsection{Effect of the kernel size $k$ in PCPG} \label{section4.2}
Different kernel size $k$ corresponds to the different receptive fields when computing the reward at each time step. In theory, choices of different size of the receptive fields should bring different effects on the final lip-reading performance. To explore the impact of different kernel size $k$, we perform several different experiments on the sentence-level benchmark GRID, because the samples in sentence-level are long enough to test the effects of different $k$. In this part, we keep $s$ at 1 to make the model have more overlapping parts. To make the pseudo-convolutional kernel put the same attention to the reward at each time step, the $k$-dimensional kernel weight $w$ is set to [$\dfrac{1}{k}$,$\dfrac{1}{k}$,\dots,$\dfrac{1}{k}$]. The results are shown in TABLE \ref{table2:a}. As is shown, we get the best result when $k=3$. When $k$ is too small (such as $k=2$), the context considered to compute the reward at each time step is not much enough and so there is an indeed improvement but not too much. When $k$ is too big (such as $k=7$), the context considered at each time step is so much that it may cover up and so weaken the contribution of the current time step. But no matter which value $k$ is, the performance is always better than the baseline $k=1$ when $k>1$. 
\vspace{-0.09cm}

\subsection{Effect of the kernel weight $w$ in PCPG}
\vspace{-0.00cm}
Different choices of the kernel weight means the different weight values would be put on the contextual time steps when computing the reward at each targeted time step. In this experiments, we fix $k$ to the above-optimized value 3 and also keep $s$ to 1 to evaluate and compare the effect of different kernel weights $w$ on the sentence-level benchmark GRID, as shown in TABLE \ref{table2:b}. From this table, we can easily see that the performance with different kernel weights is almost kept at the same level where the gap between the best and the worst is no more than $1\%$, which shows the robustness of the PCPG based seq2seq model with respect to the value of the weight.
\vspace{-0.0cm}
\subsection{Evaluation of the learned representation}
To evaluate the representation learned by the PCPG based seq2seq model, we fixed the video encoder with the same parameters as the learned PCPG based seq2seq model, and then train an FC based classifier to perform sequence-level classification back-end on the word-level lip-reading dataset LRW and LRW-1000. The results are shown in TABLE \ref{table4}. From this table, we can see that when we use the representation learned by the PCPG based seq2seq model, there is a clear improvement. And when training the representation and the FC based classifier together, the improvement is getting more obvious. We also compare with other sequence-level classification based methods in Table IV, which also clearly show the effectiveness of our method.

\subsection{Comparison with state-of-the-art  methods}
Besides the above thorough comparison and analysis of our proposed PCPG based seq2seq in different settings, we also perform a comparison with other related state-of-the-art methods, including both sentence-level and word-level methods. Please note that we have not counted in the methods using large-scale extra data except the published dataset itself for fair comparison here. As shown in TABLE \ref{table5},  we can see that our proposed methods achieve state-of-the-art performance in the decoding tasks, no matter with or without beam search (BM). As shown in  TABLE \ref{table6}, in the classifying tasks, our method has also achieved a significant improvement, especially on the LRW-1000 where the improvement is about 0.5 percents which is always hard to obtain for the difficulty of this dataset. These results clearly prove the effectiveness of the proposed PCPG module and the PCPG based seq2seq model for lip-reading.

\vspace{-0.0cm}
\begin{table}[H]
	\setlength{\abovecaptionskip}{0.1cm}
	\caption{Comparison with other classification based on methods on LRW and LRW-1000.} \label{table6}
	\centering
	\begin{tabular}{|p{1.39cm}|c|c|}    
		\hline
		Dataset & Method&  Accuracy $\uparrow$\\
		\hline
		\hline
		\multirow{5}{*}{LRW}&\cite{Petridis2018}  & 82.0$\%$  \\
		~&\cite{Stafylakis2017} & 83.0$\%$ \\
		~&\cite{themos}&{82.9}$\%$  \\
		~&\cite{Wang2019}&{83.34}$\%$  \\
		~&ours& \textbf{83.5}$\%$  \\
		\hline
		\multirow{3}{*}{LRW-1000}& \cite{Wang2019}& {36.91}$\%$    \\  
		~&    \cite{Yang2019} & 38.19$\%$ \\
		~&ours  & \textbf{38.70}$\%$    \\
		
		\hline
	\end{tabular}
	
\end{table}

\vspace{-0.15cm}
\begin{table}[H]
	\setlength{\abovecaptionskip}{0.0cm}
	\caption{Comparison with other decoding based methods on GRID, LRW, and LRW-1000. For LRW and  LRW-1000, Accuracy equals to 1$-$WER.  (The work in \cite{Afouras2017} is published in a poster from the University of Oxford.) } \label{table5}
	\centering
	\begin{subtable}[t]{2in} 
		\centering
		\begin{tabular}{|p{1.39cm}|c|c|}    
			\hline
			Dataset & Method& WER $\downarrow$ \\
			\hline
			\hline
			\multirow{6}{*}{GRID}&\cite{Wand2016} & 20.4$\%$      \\
			~&\cite{Gergen2016} & 13.6$\%$     \\
			~&\cite{Assael2016} (No BM) & 13.6$\%$         \\
			~&ours (No BM) & \textbf{11.9}$\%$        \\
			~&\cite{Assael2016} (With BM) & {11.4}$\%$       \\
			~&ours (With BM) & \textbf{11.2}$\%$     \\
			\hline
		\end{tabular}
		\vspace{0.04cm}
		\caption{Decoding results on GRID}\label{table5:(a)}
	\end{subtable}
	\begin{subtable}[t]{3in}
		\centering
		\begin{tabular}{|p{1.39cm}|c|c|}    
			\hline
			Dataset & Method& Accuracy $\uparrow$ \\
			\hline
			\multirow{2}{*}{LRW}&\cite{Afouras2017} &76.2$\%$  \\
			~&ours & \textbf{78.5}$\%$  \\    
			\hline
			\multirow{1}{*}{LRW-1000}&ours &\textbf{33.1}$\%$  \\
			\hline    
		\end{tabular}
		\vspace{0.2cm}
		\caption{Decoding results on LRW and LRW-1000}\label{table5:b}
	\end{subtable}
\end{table}

\section{Conclusion}
In this work, we proposed a pseudo-convolutional policy gradient (PCPG) based seq2seq model for the lip-reading task. Inspired by the principle of convolutional operation, we consider to extend the policy gradient's receptive field and overlapping parts in the training process. We perform a thorough evaluation of both the word-level and the sentence-level dataset. Compared with the state-of-the-art results, the PCPG outperforms or equals to the state-of-the-art performance, which verifies the advantages of the PCPG. Moreover, the PCPG can also be applied to other seq2seq tasks, such as machine translation, automatic speech recognition, image caption, video caption and so on. 

\section{ACKNOWLEDGMENTS}
This work is partially supported by National Key R\&D Program of China (No. 2017YFA0700804) and National Natural Science Foundation of China (No. 61702486, 61876171).


\bibliographystyle{ieee}
\bibliography{egbib}

\begin{thebibliography}{10}\itemsep=-1pt

\bibitem{Afouras2018}
T.~Afouras, J.~S. Chung, A.~W. Senior, O.~Vinyals, and A.~Zisserman.
\newblock Deep audio-visual speech recognition.
\newblock {\em IEEE transactions on pattern analysis and machine intelligence},
  2018.

\bibitem{Afouras2017}
T.~Afouras, J.~S. Chung, and A.~Zisserman.
\newblock Deep learning for lip reading.
\newblock {\em Poster Presentation in the University of Oxford}, 2017.

\bibitem{Afouras}
T.~Afouras, J.~S. Chung, and A.~Zisserman.
\newblock Deep lip reading: a comparison of models and an online application.
\newblock {\em In Proceedings of Interspeech}, abs/1806.06053, 2018.

\bibitem{Assael2016}
Y.~M. Assael, B.~Shillingford, S.~Whiteson, and N.~de~Freitas.
\newblock Lipnet: End-to-end sentence-level lipreading.
\newblock 2017.

\bibitem{Banerjee2003}
S.~Banerjee and A.~Lavie.
\newblock Meteor: An automatic metric for mt evaluation with improved
  correlation with human judgments.
\newblock In {\em IEEvaluation@ACL}, 2005.

\bibitem{Berkenkamp2017}
F.~Berkenkamp, M.~Turchetta, A.~P. Schoellig, and A.~Krause.
\newblock Safe model-based reinforcement learning with stability guarantees.
\newblock In {\em NIPS}, 2017.

\bibitem{Chung}
J.~S. Chung, A.~W. Senior, O.~Vinyals, and A.~Zisserman.
\newblock Lip reading sentences in the wild.
\newblock {\em 2017 IEEE Conference on Computer Vision and Pattern Recognition
  (CVPR)}, pages 3444--3453, 2016.

\bibitem{B2017}
J.~S. Chung and A.~Zisserman.
\newblock Lip reading in the wild.
\newblock In {\em ACCV}, 2016.

\bibitem{Chung2017}
J.~S. Chung and A.~Zisserman.
\newblock Lip reading in profile.
\newblock In {\em BMVC}, 2017.

\bibitem{Chung2018}
J.~S. Chung and A.~Zisserman.
\newblock Learning to lip read words by watching videos.
\newblock {\em Comput. Vis. Image Underst.}, 173:76--85, 2018.

\bibitem{cooke2006}
M.~Cooke, J.~Barker, S.~Cunningham, and X.~D. Shao.
\newblock An audio-visual corpus for speech perception and automatic speech
  recognition.
\newblock {\em The Journal of the Acoustical Society of America}, pages
  2421--4, 2006.

\bibitem{Gergen2016}
S.~Gergen, S.~Zeiler, A.~H. Abdelaziz, R.~M. Nickel, and D.~Kolossa.
\newblock Dynamic stream weighting for turbo-decoding-based audiovisual asr.
\newblock In {\em INTERSPEECH}, 2016.

\bibitem{He}
K.~He, X.~Zhang, S.~Ren, and J.~Sun.
\newblock Deep residual learning for image recognition.
\newblock {\em 2016 IEEE Conference on Computer Vision and Pattern Recognition
  (CVPR)}, pages 770--778, 2015.

\bibitem{Hu2016}
D.~Hu, X.~Li, and X.~Lu.
\newblock Temporal multimodal learning in audiovisual speech recognition.
\newblock {\em 2016 IEEE Conference on Computer Vision and Pattern Recognition
  (CVPR)}, pages 3574--3582, 2016.

\bibitem{Jin2019}
M.~Jin and J.~Lavaei.
\newblock Control-theoretic analysis of smoothness for stability-certified
  reinforcement learning.
\newblock {\em 2018 IEEE Conference on Decision and Control (CDC)}, pages
  6840--6847, 2018.

\bibitem{Lan2009}
Y.~Lan, R.~Harvey, B.-J. Theobald, E.-J. Ong, and R.~Bowden.
\newblock Comparing visual features for lipreading.
\newblock In {\em AVSP}, 2009.

\bibitem{Lin2001}
C.-Y. Lin.
\newblock Rouge: A package for automatic evaluation of summaries.
\newblock In {\em ACL 2004}, 2004.

\bibitem{Nikishin2018}
E.~Nikishin, P.~Izmailov, B.~Athiwaratkun, D.~Podoprikhin, T.~Garipov,
  P.~Shvechikov, D.~P. Vetrov, and A.~G. Wilson.
\newblock Improving stability in deep reinforcement learning with weight
  averaging.
\newblock 2018.

\bibitem{Papineni2002}
K.~Papineni, S.~Roukos, T.~Ward, and W.-J. Zhu.
\newblock Bleu: a method for automatic evaluation of machine translation.
\newblock In {\em ACL}, 2001.

\bibitem{Petridis2018}
S.~Petridis, T.~Stafylakis, P.~Ma, F.~Cai, G.~Tzimiropoulos, and M.~Pantic.
\newblock End-to-end audiovisual speech recognition.
\newblock {\em 2018 IEEE International Conference on Acoustics, Speech and
  Signal Processing (ICASSP)}, pages 6548--6552, 2018.

\bibitem{Chopra2016}
M.~Ranzato, S.~Chopra, M.~Auli, and W.~Zaremba.
\newblock Sequence level training with recurrent neural networks.
\newblock {\em CoRR}, abs/1511.06732, 2015.

\bibitem{Rennie}
S.~J. Rennie, E.~Marcheret, Y.~Mroueh, J.~Ross, and V.~Goel.
\newblock Self-critical sequence training for image captioning.
\newblock {\em 2017 IEEE Conference on Computer Vision and Pattern Recognition
  (CVPR)}, pages 1179--1195, 2016.

\bibitem{themos}
T.~Stafylakis, M.~H. Khan, and G.~Tzimiropoulos.
\newblock Pushing the boundaries of audiovisual word recognition using residual
  networks and lstms.
\newblock {\em Comput. Vis. Image Underst.}, 176-177:22--32, 2018.

\bibitem{Stafylakis2017}
T.~Stafylakis and G.~Tzimiropoulos.
\newblock Combining residual networks with lstms for lipreading.
\newblock {\em ArXiv}, abs/1703.04105, 2017.

\bibitem{Tech}
R.~Vedantam, C.~L. Zitnick, and D.~Parikh.
\newblock Cider: Consensus-based image description evaluation.
\newblock {\em 2015 IEEE Conference on Computer Vision and Pattern Recognition
  (CVPR)}, pages 4566--4575, 2014.

\bibitem{Wand2016}
M.~Wand, J.~Koutn{\'i}k, and J.~Schmidhuber.
\newblock Lipreading with long short-term memory.
\newblock {\em 2016 IEEE International Conference on Acoustics, Speech and
  Signal Processing (ICASSP)}, pages 6115--6119, 2016.

\bibitem{Wang2019}
C.~Wang.
\newblock Multi-grained spatio-temporal modeling for lip-reading.
\newblock {\em ArXiv}, abs/1908.11618, 2019.

\bibitem{Wiseman2016}
S.~Wiseman and A.~M. Rush.
\newblock Sequence-to-sequence learning as beam-search optimization.
\newblock In {\em EMNLP}, 2016.

\bibitem{Yang2019}
S.~Yang, Y.~Zhang, D.~Feng, M.~Yang, C.~Wang, J.~Xiao, K.~Long, S.~Shan, and
  X.~Chen.
\newblock Lrw-1000: A naturally-distributed large-scale benchmark for lip
  reading in the wild.
\newblock {\em 2019 14th IEEE International Conference on Automatic Face \&
  Gesture Recognition (FG 2019)}, pages 1--8, 2018.

\bibitem{Zhou2014}
Z.~Zhou, G.~Zhao, X.~Hong, and M.~Pietik{\"a}inen.
\newblock A review of recent advances in visual speech decoding.
\newblock {\em Image Vis. Comput.}, 32:590--605, 2014.

\end{thebibliography}

\end{document}